\documentclass{article}

\usepackage{PRIMEarxiv}

\usepackage[utf8]{inputenc} 
\usepackage[T1]{fontenc}    
\usepackage{hyperref}       
\usepackage{url}            
\usepackage{booktabs}       
\usepackage{amsfonts}       
\usepackage{nicefrac}       
\usepackage{microtype}      
\usepackage{lipsum}
\usepackage{fancyhdr}       
\usepackage{graphicx}       
\graphicspath{{media/}}     
\usepackage{amsmath}
\usepackage{amssymb}
\usepackage{xcolor}  

\title{Ordinary Least Squares is a Special Case of Transformer
}

\author{
  Xiaojun Tan \\
  Center for Peak Performance, Department of Psychology and Behavioral Sciences, Zhejiang University\\
  Hangzhou Higgs Asset Management Co., Ltd. \\
  \texttt{tanxiaojun@higgsasset.com}
  \And
  Yuchen Zhao \\
  College of Control Science and Engineering, Zhejiang University \\
  Hangzhou Higgs Asset Management Co., Ltd. \\
  \texttt{zhaoyuchen@zju.edu.cn}
}

\begin{document}
\maketitle

\begin{abstract}
The statistical essence of the Transformer architecture has long remained elusive: Is it a universal approximator, or a neural network version of known computational algorithms? Through rigorous algebraic proof, we show that the latter better describes Transformer's basic nature: Ordinary Least Squares (OLS) is a special case of the single-layer Linear Transformer. Using the spectral decomposition of the empirical covariance matrix, we construct a specific parameter setting where the attention mechanism’s forward pass becomes mathematically equivalent to the OLS closed-form projection. This means attention can solve the problem in one forward pass, not by iterating. Building upon this prototypical case, we further uncover a decoupled slow and fast memory mechanism within Transformers. Finally, the evolution from our established linear prototype to standard Transformers is discussed. This progression facilitates the transition of the Hopfield energy function from linear to exponential memory capacity, thereby establishing a clear continuity between modern deep architectures and classical statistical inference.
\end{abstract}


\section{Introduction}
\label{sec: introduction}

The Transformer architecture has become the solid standard for processing large-scale data~\cite{nips2017AttentionisAllYouNeed}. However, despite its success in engineering applications, a unified and concrete explanation regarding its mathematical essence and its links to classical statistical computation remains elusive. Current research has proposed several theoretical approaches. For instance, some studies view the Transformer architecture through associative memory in modern Hopfield networks~\cite{arxiv2021HopfieldisAllYouNeed}. Others analyze it as an evolution of particle system dynamics~\cite{arxiv2026MeanField}, clustering behavior~\cite{nips2023TransformerCluster} or a Bayesian inference framework~\cite{arxiv2026TransformerBayesian}. While these explanations provide high-level theoretical insights, they are often too abstract to directly characterize the model's precise operation behavior when executing fundamental statistical tasks during a forward pass.

To bridge the gap between abstract theory and concrete algebraic constructions, we examine the Transformer architecture using the most basic statistical inference task: linear regression. Recent studies have revealed a remarkable capability of the Transformer to solve linear regression problems without task-specific parameter updates~\cite{nips2024multilayerTransformerRegression}. This capability is commonly interpreted as a simulation of iterative optimization processes, like traditional statistical learning models~\cite{nips2023TransformerICLProve}. Such perspectives suggest that the architecture implicitly executes gradient descent to gradually approach the optimal solution~\cite{icml2023TransformerICLGD,nips2023TransformerGDICL}. Such interpretations treat the Transformer fundamentally as an asymptotic approximator, emphasizing its multi-step iterative refinement and learning-rate-dependent convergence process~\cite{jmlr2024TrainedTransformerOLS}.

In contrast, this paper introduces a more fundamental perspective. We propose and rigorously prove that the relationship is far more direct than currently recognized: Ordinary Least Squares (OLS) regression is structurally a special case of a single-layer Linear Transformer. Specifically, spectral decomposition reveals that linear attention achieves the OLS projection in one forward pass, showing statistical inference to be Transformer's intrinsic algebraic property rather than an iterative simulation. This prototype further exposes a decoupled slow-fast memory mechanism and establishes a theoretical foundation for Hopfield associative memory. The contributions of this paper are summarized as follows:

\textbf{1. Structural isomorphism between Transformer and OLS.} We establish the formal equivalence between a single-layer Linear Transformer and the OLS estimator through a theoretical proof. By introducing the OLS-Transformer as a specialized case, we empirically validate its functional convergence and structural emergence toward the analytical OLS solution during the training process.

\textbf{2. Transformer’s memory decoupling mechanism.} Based on the OLS-Transformer, we propose a slow and fast memory framework: the weight matrices serve as slow memory to extract long-term statistical patterns, while the attention scores act as fast memory to construct real-time contextual associations. This perspective provides a theoretical foundation for analyzing the model's generalization and context-aware behavior.

\textbf{3. Evolution and associative memory in Transformers.} Taking the OLS-Transformer as a starting point, we explore five dimensions of its evolution toward the standard Transformer. Notably, we point out that the transition from linear projection to Softmax attention represents a fundamental leap in the energy function of Hopfield associative memory networks, ultimately achieving a breakthrough in memory capacity from linear to exponential scale.

The rest of this paper is structured to detail these contributions in Sections \ref{sec: main theorem}, \ref{sec: slow and fast memory} and \ref{sec: ols to transformer}, respectively, and the final conclusions are made in Section \ref{sec: conclusion}.

\section{Main Theorem: OLS as a Special Case of Linear Transformer}
\label{sec: main theorem}

\subsection{Theorem Statement}
\label{subsec: theorem statement}

Consider a dataset consisting of a design matrix $\mathbf{X} \in \mathbb{R}^{n \times k}$ and a response vector $\mathbf{Y} \in \mathbb{R}^{n \times 1}$. We assume $\mathbf{X}$ is of full column rank. The goal of OLS is to find a coefficient vector $\boldsymbol{\beta}$ that minimizes the squared error $\|\mathbf{Y} - \mathbf{X}\boldsymbol{\beta}\|^2$, yielding the closed-form prediction:
\begin{equation}
\hat{\mathbf{Y}} = \mathbf{X}\boldsymbol{\beta} = \mathbf{X}(\mathbf{X}^\text{T}\mathbf{X})^{-1}\mathbf{X}^\text{T}\mathbf{Y}.
\end{equation}

We define a single-layer Linear Transforme as a modular architecture consisting of a linear attention mechanism, a feed-forward network (FFN), and an output head. Given an input $\mathbf{X}$, the forward pass is computed as:
\begin{equation}
\text{Output}(\mathbf{X}) = \frac{1}{n} \underbrace{(\mathbf{X}\mathbf{W}_\text{Q})(\mathbf{X}\mathbf{W}_\text{K})^\text{T}(\mathbf{X}\mathbf{W}_\text{V})}_{\text{Linear Attention}} \cdot \underbrace{\mathbf{W}_\text{FFN}}_{\text{FFN}} \cdot \underbrace{\mathbf{W}_\text{P}}_{\text{Head}},
\end{equation}
where $\mathbf{W}_\text{Q}, \mathbf{W}_\text{K}, \mathbf{W}_\text{V} \in \mathbb{R}^{k \times k}$ are projection matrices, $\mathbf{W}_\text{FFN} \in \mathbb{R}^{k \times k}$ represents the feed-forward weights, and $\mathbf{W}_\text{P} \in \mathbb{R}^{k \times 1}$ is the final output projection. The factor $1/n$ serves as a normalization constant ensuring the attention energy remains invariant to the sequence length $n$. Note that we omit the Softmax activation and standard normalization layers to focus on the structural isomorphism of the linear operations.

Based on the definitions above, we assert that a single-layer Linear Transformer can exactly represent the OLS solution. To be specific, for any design matrix $\mathbf{X}$ and response $\mathbf{Y}$, there exists a parameter configuration $\{\mathbf{W}_\text{Q}, \mathbf{W}_\text{K}, \mathbf{W}_\text{V}, \mathbf{W}_\text{FFN}, \mathbf{W}_\text{P}\}$ such that the Transformer's forward pass is equivalent to the OLS projection $\hat{\mathbf{Y}} = \mathbf{X}(\mathbf{X}^\text{T}\mathbf{X})^{-1}\mathbf{X}^\text{T}\mathbf{Y}$.

\subsection{Theoretical Proof}
\label{subsec: theoretical proof}

\textbf{Spectral decomposition of the covariance matrix.} To establish the connection between the OLS estimator and Transformer architecture, we first decompose the inverse covariance structure. Let $\frac{1}{n}\mathbf{X}^\text{T}\mathbf{X} = \mathbf{V}\boldsymbol{\Lambda}\mathbf{V}^\text{T}$ be the eigen-decomposition of the empirical covariance matrix, where $\mathbf{V} \in \mathbb{R}^{k \times k}$ is an orthogonal matrix of eigenvectors and $\boldsymbol{\Lambda} = \text{diag}(\lambda_1, \ldots, \lambda_k)$ is the diagonal matrix of eigenvalues. We define an intermediate transformation matrix $\mathbf{L}$ as:
\begin{equation}
\mathbf{L} = \mathbf{V}\boldsymbol{\Lambda}^{-1/2}.
\end{equation}

By construction, the inverse of the covariance matrix can be expressed as:
\begin{equation}
(\mathbf{X}^\text{T}\mathbf{X})^{-1} = \frac{1}{n} (\mathbf{V}\boldsymbol{\Lambda}\mathbf{V}^\text{T})^{-1} = \frac{1}{n} \mathbf{V}\boldsymbol{\Lambda}^{-1}\mathbf{V}^\text{T} = \frac{1}{n} \mathbf{L}\mathbf{L}^\text{T}.
\end{equation}

Substituting this decomposition into the OLS closed-form solution, we obtain:
\begin{equation}
\hat{\mathbf{Y}} = \frac{1}{n} \mathbf{X}(\mathbf{L}\mathbf{L}^\text{T})\mathbf{X}^\text{T}\mathbf{Y} = \frac{1}{n} (\mathbf{X}\mathbf{L}) \cdot (\mathbf{L}^\text{T}\mathbf{X}^\text{T}) \cdot \mathbf{Y}.
\end{equation}

\textbf{Decomposition of the coefficient vector.} While the above equation's form suggests a direct attention-like computation where $\mathbf{X}\mathbf{L}$ serves as Query/Key and $\mathbf{Y}$ as Value, standard Transformer implementations require the attention mechanism to operate on homogeneous input sequences with task-specific information encoded in learnable projection weights. To satisfy this architectural constraint, utilize the identity $(\mathbf{X}^\text{T}\mathbf{X})^{-1} = \frac{1}{n} \mathbf{L}\mathbf{L}^\text{T}$, the OLS coefficient vector $\boldsymbol{\beta} = (\mathbf{X}^\text{T}\mathbf{X})^{-1}\mathbf{X}^\text{T}\mathbf{Y}$ further decomposes as:
\begin{equation}
\boldsymbol{\beta} = \mathbf{L}\mathbf{P}, \quad \text{where} \quad \mathbf{P} = \frac{1}{n} \mathbf{L}^\text{T} \mathbf{X}^\text{T} \mathbf{Y}.
\end{equation}
This expresses $\boldsymbol{\beta}$ in the basis defined by $\mathbf{L}$, with $\mathbf{P}$ serving as the coordinate vector. Consequently, the OLS prediction can be rewritten as $\hat{\mathbf{Y}} = \mathbf{X}\boldsymbol{\beta} = \mathbf{X}\mathbf{L}\mathbf{P}$.

\textbf{Isomorphism to the Linear Transformer architecture.} To align with the modular structure of the Transformer, we strategically decompose $\hat{\mathbf{Y}} = \mathbf{X}\mathbf{L}\mathbf{P}$ by inserting the identity $\frac{1}{n} \mathbf{L}^\text{T}\mathbf{X}^\text{T}\mathbf{X}\mathbf{L} = \mathbf{I}$ between the factors $\mathbf{X}\mathbf{L}$ and $\mathbf{P}$:
\begin{equation}
\hat{\mathbf{Y}} = (\mathbf{X}\mathbf{L}) \cdot (\frac{1}{n} \mathbf{L}^\text{T}\mathbf{X}^\text{T}\mathbf{X}\mathbf{L}) \cdot \mathbf{P} = \frac{1}{n} \underbrace{(\mathbf{X}\mathbf{L})}_{\mathbf{X}\mathbf{W}_\text{Q}} \cdot \underbrace{(\mathbf{X}\mathbf{L})^\text{T}}_{(\mathbf{X}\mathbf{W}_\text{K})^\text{T}} \cdot \underbrace{(\mathbf{X}\mathbf{L})}_{\mathbf{X}\mathbf{W}_\text{V}} \cdot \underbrace{\mathbf{I}}_{\mathbf{W}_\text{FFN}} \cdot \underbrace{\mathbf{P}}_{\mathbf{W}_{P}}.
\end{equation}

This alignment explicitly constructs the required parameter configuration: $\mathbf{W}_\text{Q} = \mathbf{L}$, $\mathbf{W}_\text{K} = \mathbf{L}$, $\mathbf{W}_\text{V} = \mathbf{L}$, $\mathbf{W}_\text{FFN} = \mathbf{I}$, and $\mathbf{W}_\text{P} = \mathbf{P}$. Under this configuration, the linear attention mechanism computes $\frac{1}{n}(\mathbf{X}\mathbf{W}_\text{Q})(\mathbf{X}\mathbf{W}_\text{K})^\text{T}(\mathbf{X}\mathbf{W}_\text{V}) = \frac{1}{n}(\mathbf{X}\mathbf{L})(\mathbf{L}^\text{T}\mathbf{X}^\text{T})(\mathbf{X}\mathbf{L})$. Grouping these operations according to the Transformer's modular hierarchy reveals the functional alignment between the statistical solver and the neural architecture:
\begin{equation}
\hat{\mathbf{Y}} = \underbrace{\frac{1}{n} (\mathbf{X}\mathbf{L}) \cdot (\mathbf{L}^\text{T}\mathbf{X}^\text{T}) \cdot (\mathbf{X}\mathbf{L})}_{\text{Linear Attention}} \cdot \underbrace{\mathbf{I}}_{\text{FFN}} \cdot \underbrace{\mathbf{P}}_{\text{Head}} = \text{Output}(\mathbf{X}).
\end{equation}

This confirms that the structure of a Linear Transformer is mathematically isomorphic to the closed-form OLS estimator. We refer to this specific configuration as the \emph{OLS-Transformer} to distinguish its dynamic in-context behavior from the static projection of traditional OLS.

\subsection{Empirical Validation}
\label{subsec: empirical validation}

The structural isomorphism established in Section~\ref{subsec: theoretical proof} provides a static parameter configuration that matches a Linear Transformer to the OLS estimator. To further investigate whether this configuration is a reachable and stable endpoint under standard gradient-based optimization, we perform an empirical validation on the convergence dynamics of the OLS-Transformer.

\textbf{Experimental setup.} We consider a one-dimensional linear regression task $y = 2x + \epsilon$, with $n=500$ samples and noise $\epsilon \sim \mathcal{N}(0, 10^{-4})$. The OLS-Transformer is implemented to strictly follow the reformulated analytical expression derived in Section~\ref{subsec: theoretical proof}: the projection matrices are constrained as $\mathbf{W}_\text{Q} = \mathbf{W}_\text{K} = \mathbf{W}_\text{V} = L$, and the task-specific weight is defined as $\mathbf{P} = \frac{1}{n} L \mathbf{X}^\text{T} \mathbf{Y}$. By reducing the learnable parameters to a single scalar $L$, we can directly monitor the trajectory of the attention parameter relative to the theoretical optimum $L^* = (\frac{1}{n}\mathbf{X}^\text{T}\mathbf{X})^{-1/2}$. The model is initialized at $L_0 = 0.5$ and optimized using the Adam algorithm for 5000 epochs.

\begin{figure*}[t]
    \centering
    \includegraphics[width=0.95\linewidth]{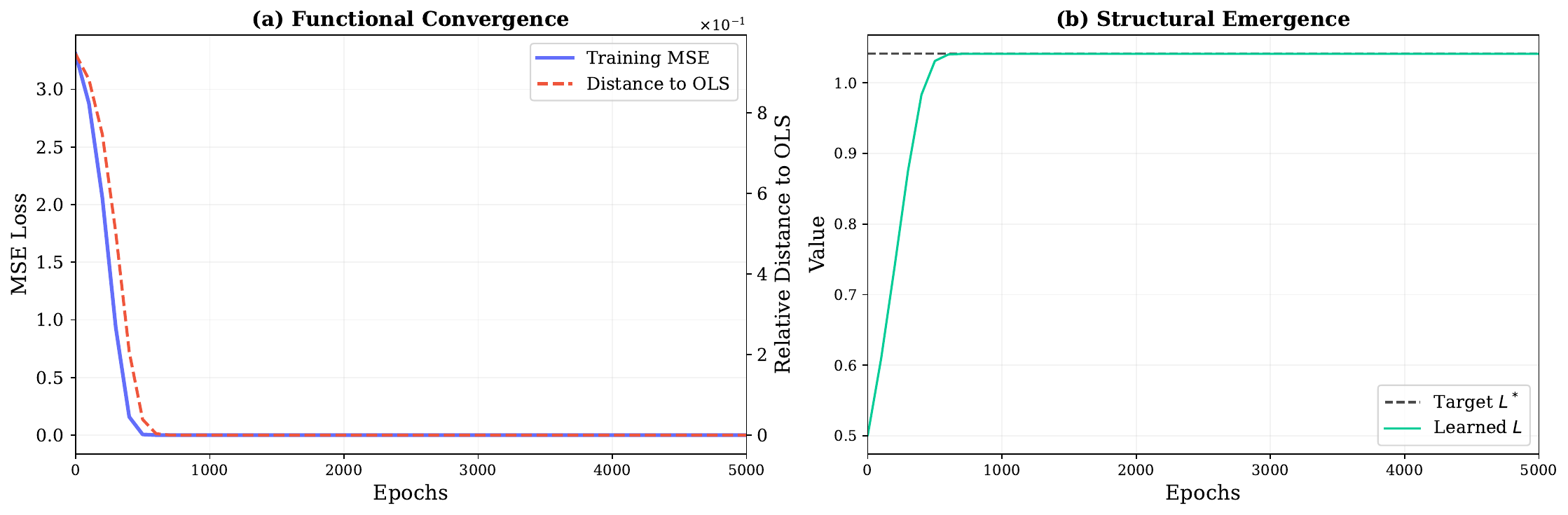}
    \caption{Empirical validation of training reachability for the OLS-Transformer. 
    \textbf{(a) Functional Convergence}: The synchronized evolution of training MSE (left axis) and relative distance to the OLS benchmark (right axis), with rapid alignment occurring within the initial 1000 epochs. 
    \textbf{(b) Structural Emergence}: The trajectory of the learnable parameter $L$ precisely recovering the theoretical optimum $L^* = (\frac{1}{n}\mathbf{X}^\text{T}\mathbf{X})^{-1/2}$, which is temporally synchronized with the functional convergence of the output.}
    \label{fig: structural_emergence}
\end{figure*}

\textbf{Functional convergence.} Figure~\ref{fig: structural_emergence} (a) illustrates the evolution of the OLS-Transformer's output behavior. The training MSE and the relative distance between the model's prediction and the OLS closed-form solution exhibit a synchronized monotonic decay. Notably, the functional alignment occurs rapidly within the first 1000 epochs, at which point the MSE stabilizes at the noise floor and the relative distance to the OLS benchmark approaches zero. This indicates that the gradient descent process effectively minimizes the objective by driving the OLS-Transformer to spontaneously execute the OLS projection logic.

\textbf{Structural emergence.} Figure~\ref{fig: structural_emergence} (b) illustrates the internal parameter dynamics, where the emergence of the OLS structure is characterized by the learnable scalar $L$ converging precisely to the theoretical value $L^*$. Unlike over-parameterized models where weight redundancy often masks the underlying evolution, our constrained OLS-Transformer explicitly recovers the exact inverse covariance structure required for OLS. Notably, this structural emergence is temporally synchronized with the functional convergence of the model's output. These results confirm that the parameter configuration of the OLS-Transformer can stably converge to the analytical solution state required by the OLS problem, proving that standard optimization can recover the exact statistical projection operator $\mathbf{X}(\mathbf{X}^\text{T}\mathbf{X})^{-1}\mathbf{X}^\text{T}$ within the linear attention framework.

\section{Slow and Fast Memory in OLS-Transformer}
\label{sec: slow and fast memory}

After reformulating OLS as a Linear Transformer architecture, a profound insight is that OLS-Transformer naturally exhibits the decoupled memory characteristics.

In artificial neural networks, weights obtained through backpropagation training are regarded as \emph{slow memory}, which define model's long-term logic for interpreting information and do not change within inference scenarios. In contrast, associations dynamically constructed during the inference stage through the attention mechanism are regarded as \emph{fast memory}, which evolve with the context in real-time, similar to certain features in Hebbian learning~\cite{nips2016FastWeights, jmlr2021FastWeights}. In OLS-Transformer, the matrix $\mathbf{L}$ represents slow memory as a fixed parameter obtained during training, while the attention score term dynamically constructed from input data belongs to the fast memory.

To further illustrate OLS-Transformer's sensitivity under these memory characteristics, we consider a training dataset $\{\mathbf{X}, \mathbf{Y}_x\}$ with sample size $n$ and an in-context inference dataset $\{\mathbf{Z}, \mathbf{Y}_z\}$ with sample size $m$. We assume both datasets share the same underlying linear relationship: $\mathbf{Y}_x = \mathbf{X}\boldsymbol{\beta}$ and $\mathbf{Y}_z = \mathbf{Z}\boldsymbol{\beta}$, where $\boldsymbol{\beta}$ is the ground-truth coefficient vector. We also define their empirical covariance matrices as $\Sigma_x = \frac{1}{n}\mathbf{X}^\text{T}\mathbf{X}$ and $\Sigma_z = \frac{1}{m}\mathbf{Z}^\text{T}\mathbf{Z}$, respectively.

From the perspective of traditional OLS, the predictor performs inference using a static logic $\hat{\boldsymbol{\beta}} = \Sigma_x^{-1} (\frac{1}{n}\mathbf{X}^\text{T}\mathbf{Y}_x)$ solidified from the training set. This logic is entirely fixed and does not change with the distribution $\Sigma_z$ of the inference set. In contrast, OLS-Transformer uses the following formula in the inference scenario $\mathbf{Z}$:
\begin{equation}
\hat{\mathbf{Y}}_z = \frac{1}{m} \mathbf{Z}\mathbf{L}\mathbf{L}^\text{T}\mathbf{Z}^\text{T}\mathbf{Y}_z.
\end{equation}
where $\mathbf{L}\mathbf{L}^\text{T} = \Sigma_x^{-1}$ is the slow memory learned based on the training set distribution, while $\frac{1}{m}\mathbf{Z}^\text{T}\mathbf{Y}_z$ is the fast memory constructed based on the current inference context. Substituting the underlying relationship $\mathbf{Y}_z = \mathbf{Z}\boldsymbol{\beta}$ yields:
\begin{equation}
\hat{\mathbf{Y}}_z = \mathbf{Z} \left[ \Sigma_x^{-1} \Sigma_z \right] \boldsymbol{\beta}.
\end{equation}

This derivation clearly demonstrates OLS-Transformer's sensitivity to data distribution changes. Specifically, when the inference environment's covariance structure is consistent with that of the training environment, the middle term $\Sigma_x^{-1} \Sigma_z = \mathbf{I}$, and the model accurately recovers the OLS analytical solution $\mathbf{Z}\boldsymbol{\beta}$. However, when the distribution of $\mathbf{Z}$ deviates from $\mathbf{X}$, the prediction results will be linearly distorted by the term $\Sigma_x^{-1} \Sigma_z$. This means that OLS-Transformer can perceive the current context in real-time, but it also makes its prediction logic highly susceptible to interference from distributional inconsistencies.

\section{From OLS-Transformer to Standard Transformer}
\label{sec: ols to transformer}

While fast memory enables context-aware predictions, OLS-Transformer is sensitive to changes in the data distribution, which imposes higher demands on the stability and expressivity of slow memory. To maintain high context sensitivity while enhancing inference robustness under complex distributional shifts, industrial-grade standard Transformer models must evolve beyond the simple linear association pattern of OLS-Transformer. Compared to OLS-Transformer, standard Transformer models have undergone at least the following key modifications:

\textbf{1. Nonlinear activation functions.} Standard Transformers include nonlinear activation functions to extend the model from linear regression to nonlinear feature mapping.

\textbf{2. QKV parameterization.} Standard Transformers use distinct matrices $\mathbf{W}_\text{Q}$, $\mathbf{W}_\text{K}$, and $\mathbf{W}_\text{V}$ instead of a single matrix $\mathbf{L}$, allowing the model to represent asymmetric associations between features.

\textbf{3. Softmax transformation.} Standard Transformers apply the Softmax operator to attention scores, which transforms the linear geometric projection into a probabilistic weighting of inputs.

\textbf{4. Multi-head attention.} Standard Transformers use multi-head attention to process different feature subspaces in parallel, which improves the model's generalization ability through a form of ensemble learning.

\textbf{5. Positional encoding.} Standard Transformers add positional encodings to the input, enabling the model to represent sequential information instead of being permutation invariant.

Among the many improvements mentioned above, the feature of the standard Transformer that has gained the most research attention is the Softmax attention mechanism, which also leads to several profound insights. These studies have increasingly revealed the connection between the Transformer's attention mechanism, Hopfield networks, and human cognitive processes~\cite{cell2020Tolman, naturereview2022BrainAttractor, arxiv2022TransformerNeuralRep}. In 1982, Hopfield proposed the earliest algorithm for implementing associative memory, known as the Hopfield network. However, the memory capacity of the original Hopfield network was quite limited: a system with $d$ neurons taking values of $\pm 1$ could only store approximately $0.15d$ distinct patterns~\cite{pnas1982Hopfield}.

Later, Krotov and Hopfield proposed that modifying the form of the energy function in a Hopfield network could enable higher-density pattern memory. The original Hopfield network can be viewed as a quadratic function of the inner product between the query vector and the key patterns. By changing this function to a higher-order power function, such as a polynomial of degree $k$, the memory capacity of the Hopfield network can be increased to the order of $d^{k-1}$~\cite{nips2016DenseAssociativeMemory}. Demircigil et al. further transitioned the energy function from a power function to an exponential form, allowing the memory capacity to reach the order of $2^{d/2}$~\cite{jsp2017MemoryCapacity}. Ramsauer et al.~\cite{arxiv2021HopfieldisAllYouNeed} pointed out that a slight modification to this exponential energy function, along with extending the neuron states from discrete $\pm 1$ values to continuous states, yielding the Softmax attention mechanism used in Transformers. 

In the OLS-Transformer, the attention mechanism is essentially the original Hopfield network generalized to continuous states. Therefore, the evolution from the attention pattern of the OLS-Transformer to that of the standard Transformer can be viewed as an evolution of the storage and retrieval mechanisms within associative memory.

\section{Conclusion}
\label{sec: conclusion}

By establishing a constructive isomorphism between a single-layer Linear Transformer and the OLS estimator, we prove that OLS is fundamentally a special case of a single-layer Linear Transformer under specific parameter configurations. The profound significance of this finding lies in its reduction of the Transformer from an uninterpretable ``black-box'' approximator to a statistical operator with a well-defined analytical target, demonstrating its complex behaviors can be decomposed into fundamental algebraic operations. Through the OLS-Transformer as an interpretable benchmark, we further reveal that Transformer's context-awareness lies in its decoupling mechanism between slow and fast memory. Furthermore, we elucidate its linear associative memory capacity as a quadratic Hopfield prototype and discuss the evolutionary necessity of transitioning toward exponential energy functions represented by Softmax attention.

In fact, our algebraic approach provides a critical theoretical basis for understanding the emergent behaviors of complex architectures. It strongly suggests that the remarkable context-aware reasoning and memory capacities of modern LLMs are deeply rooted in basic statistical operators with exact analytical solutions such as OLS. However, the high sensitivity of the OLS-Transformer to data distribution shifts also implies that modern designs including nonlinear activations, Softmax transformation, multi-head attention, and multi-layer stacking are essential evolutions to handle this sensitivity and enhance robustness while preserving Transformer's context-awareness.

Building upon the theoretical framework developed in this work, we can systematically explore novel operators beyond current linear attention mechanisms, such as extensions to higher-order polynomial or exponential energy functions, to reach a better balance between computational efficiency, memory density, and model interpretability. Ultimately, this will establish a solid algebraic foundation for designing the next generation of general AI architectures.

\bibliographystyle{unsrt}  
\bibliography{references}

\end{document}